# CONTOUR-BASED 3D TONGUE MOTION VISUALIZATION USING ULTRASOUND IMAGE SEQUENCES

*Kele Xu[1,2], Yin Yang[3], Clémence Leboullenger[1,2], Pierre Roussel[2], Bruce Denby[4*]*

[1]Université Pierre et Marie Curie; Paris, 75005, France
[2]Langevin Institute, ESPCI-ParisTech, Paris, 75005, France
[3]University of New Mexico, Albuquerque, New Mexico, 87102, U.S.A
[4]Tianjin University, Tianjin, 300000 China

`kelele.xu@gmail.com, yangy@unm.edu, clemence.leboullenger@gmail.com, pierre.roussel@espci.fr, denby@ieee.org`

## ABSTRACT

This article describes a contour-based 3D tongue deformation visualization framework using B-mode ultrasound image sequences. A robust, automatic tracking algorithm characterizes tongue motion via a contour, which is then used to drive a generic 3D Finite Element Model (FEM). A novel contour-based 3D dynamic modeling method is presented. Modal reduction and modal warping techniques are applied to model the deformation of the tongue physically and efficiently. This work can be helpful in a variety of fields, such as speech production, silent speech recognition, articulation training, speech disorder study, etc.

## 1. INTRODUCTION

In speech production research, realistic 3D tongue motion visualization is of importance and an accurately quantified description of the 3D tongue motion may also be helpful for a Silent Speech Interface (SSI) system [1], which employs different sensors to capture non-acoustic features for speech recognition and synthesis. Furthermore, 3D dynamic tongue modeling can serve as a tool to study articulation training [2]. However, despite considerable efforts, "seeing speech", as the process is often defined, remains a challenge.

B-mode ultrasound imaging is widely used to visualize the motion of the tongue, and is non-invasive and easy to implement. Furthermore, advances in physics-based 3D modeling technique have advanced the technique to a point where ultrasound-based 3D tongue modeling may today be feasible. In this paper, we explore a novel tongue visualization framework, which combines the 2D ultrasound imaging and a contour-based 3D physics-based modeling technique. Contours are extracted from the ultrasound tongue image sequence, and then used to drive the deformation of a 3D tongue model.

Different approaches can be proposed to follow the motion of the tongue in the ultrasound image sequences, which can be divided into two main types of methods: speckle tracking and contour tracking. The classical methods to track speckle include optical-flow and block-matching [3]. In an earlier work [4], the performance of the speckle tracking was found to be somewhat unstable. Compared to speckle tracking, the extraction of the contour of tongue surface from ultrasound images exhibits superior performance and robustness. In this paper, using contours extracted from the 2D image, we explore a novel 3D dynamic framework to model the tongue motion in a dynamic way.

The article aims to give an overall technical description of this framework, based on which a platform has been developed. The organization of the paper is as follows: In section 2, a description of the relation to prior work is given. The technical details of the contour-based 3D motion visualization are given in section 3 and section 4. Results are presented in section 5, and section 6 provides conclusion and discusses future work.

## 2. RELATION TO PRIOR WORK

In this section we discuss the relation to prior work, including contour tracking and 3D tongue modeling. For contour tracking in ultrasound tongue image sequences, many algorithms have been proposed. Previous work can be briefly divided into three kinds of approaches: active contour models [5], [6], [7]; machine learning-based tracking [8], [9] and ultrasound image segmentation-based approaches [10]. Contour tracking still has difficulties in

---

different imaging situations for different subjects. Missing contours may occur when the tongue surface is parallel to the propagation direction of the ultrasound wave, but this is outside the scope of the present paper; here, the goal is to explore a method to use extracted contours to drive the 3D tongue model.

Most previous work on modeling the dynamic 3D tongue has focused on muscle driven activation [11], [12], [13], and [14], or geometry data-driven method [15]. However, our understanding of bio-mechanical property of the human tongue is still very limited. Rather than muscle-driven 3D tongue modeling, motion-derived 3D modeling is used in our framework, as an alternate type of dynamic tongue modeling. Furthermore, the use of modal reduction and modal warping techniques allows real-time tongue visualization.

Here the extracted contour is used to drive the motion of the tongue, which can generate a more realistic simulation, since, as mentioned in the introduction (see also [2]), speckle tracking may fail when the contour disappears, giving a non-physical deformation of the tongue.

## 3. PHYSICS-BASED 3D TONGUE MODELING

The dynamics of an input tongue shape, discretized using a finite element mesh can be expressed as:
$$\mathbf{M\ddot{u}} + \mathbf{C\dot{u}} + \mathbf{Ku} = \mathbf{f} \quad (1)$$
where $\mathbf{M}$, $\mathbf{C}$, $\mathbf{K}$ are the mass, damping, and stiffness matrices, respectively, of size $3n \times 3n$ ($n$ is the number of nodes on the mesh); $\mathbf{u}$ is the vector storing the displacements of all nodes from their initial positions; and $\mathbf{f}$ is the vector of external forces. To speed up the solution of this ordinary differential equation (ODE), linear modal analysis is used. Suppose $\mathbf{\Phi}$ (whose column vectors are eigenvectors) and $\mathbf{\Lambda}$ (a diagonal matrix of eigenvalues) are solutions to a generalized eigen-problem $\mathbf{K\Phi} = \mathbf{M\Phi\Lambda}$, such that $\mathbf{\Phi}^T\mathbf{M\Phi} = \mathbf{I}$ and $\mathbf{\Phi}^T\mathbf{K\Phi} = \mathbf{\Lambda}$. We can use a linear combination of the columns in $\mathbf{\Phi}$ to express $\mathbf{u}$ as:
$$\mathbf{u} = \mathbf{\Phi q} \quad (2)$$
Here, we may take only a few dominant columns in $\mathbf{\Phi}$, which are associated with eigenvectors of small eigenvalues, thus significantly reducing the computation intensity. Substituting (2) into (1) followed by a pre-multiplication of $\mathbf{\Phi}^T$ decouples (1) as:
$$\mathbf{M}_q\ddot{\mathbf{q}} + \mathbf{C}_q\dot{\mathbf{q}} + \mathbf{K}_q\mathbf{q} = \mathbf{\Phi}^T\mathbf{f} \quad (3)$$
where $\mathbf{M}_q = \mathbf{I}$, $\mathbf{C}_q = \xi\mathbf{I} + \zeta\mathbf{\Lambda}$ ($\xi$ and $\zeta$ are scalar weighting factors of the damping), $\mathbf{K}_q = \mathbf{\Lambda}$. $\mathbf{\Phi}^T\mathbf{f}$ is called the modal force. It is to be noted that $\mathbf{M}_q$, $\mathbf{C}_q$ and $\mathbf{K}_q$ are now all diagonal matrices of a much smaller size, and the time integration for (3) can be carried out in real-time.

We also use the modal warping technique [16] to compute the nonlinear deformation term, so that large rotational deformations of the tongue can be well captured.

More detailed formulation and derivation can be found in [2], [17].

## 4. CONTOUR-BASED 3D TONGUE MOTION VISUALIZATION

The 3D model can be driven by imposing extra positional constraints at specified finite element nodes to enforce their displacements to some user specified values. To drive the 3D tongue model, the modal displacement needs to be calculated by making use of the contour extracted from the ultrasound image sequences. However, obtaining the correspondence between contours of different frames is of great difficulty, and registration between the 2D ultrasound image and 3D tongue model a major challenge. Rather than using speckle tracking, in this paper, we show that these challenges can actually be converted into a "3D shape search" problem. The detailed method is given as follows:

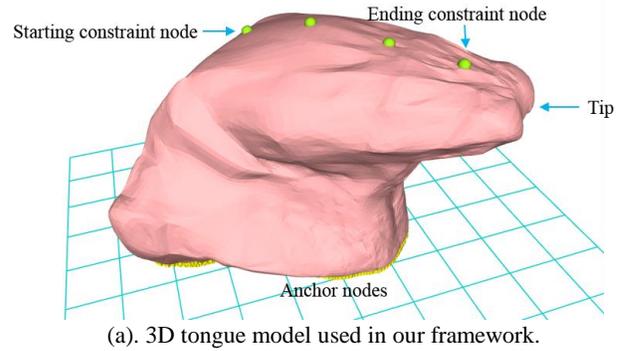

(a). 3D tongue model used in our framework.

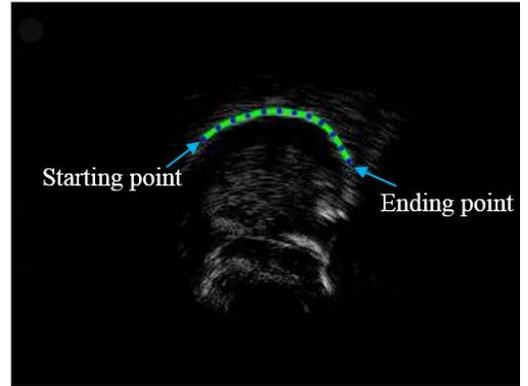

(b). Ultrasound tongue image with contour extracted.

**Fig. 1.** Elements used for the 3D visualization. (a) The 3D model used in our framework, the green circles denote the constraint nodes, whose displacements are associated with the modal displacement. The yellow nodes are anchor nodes whose displacements are zero during the deformation of the tongue model. (b) Target curve extracted from the image, the green lines are the surface of the tongue.

**Step 1: Initialization.** Four constraint nodes are selected manually (as shown in Fig. 1(a)). In this paper, we suppose the first and last nodes are associated to the starting points and ending points of the contour extracted from the 2D image (as shown in Fig. 1(b)).

**Step 2: Database Construction.** Each constraint node on the 3D tongue model has 2 degrees of freedom. At each time step, a constraint point will be assigned random displacements along the X-axis and Y-axis in the midsagittal plane. Because the movement of the tongue is smooth, we set up an upper threshold to the magnitude of the displacement so as to eliminate any discontinuous deformations. The 3D tongue model will then generate different tongue shapes, which are used to construct a 3D tongue shape database (some samples from the dataset are given in Fig. 2.). As the displacement is random, some unphysical 3D tongue shapes will be generated, which will be discarded manually. For every 3D tongue shape in the database, a contour can be extracted from the model by using the nodes lying on the surface between the starting node and ending node in the mid-sagittal plane. As the movement of the tongue can be viewed as symmetric, the 3D contours from the database can be projected into the mid-sagittal 2D plane, and compared to the target curve extracted from the 2D ultrasound image. In our experiment, the number of 3D sample tongue shapes in the database is 1000 presently.

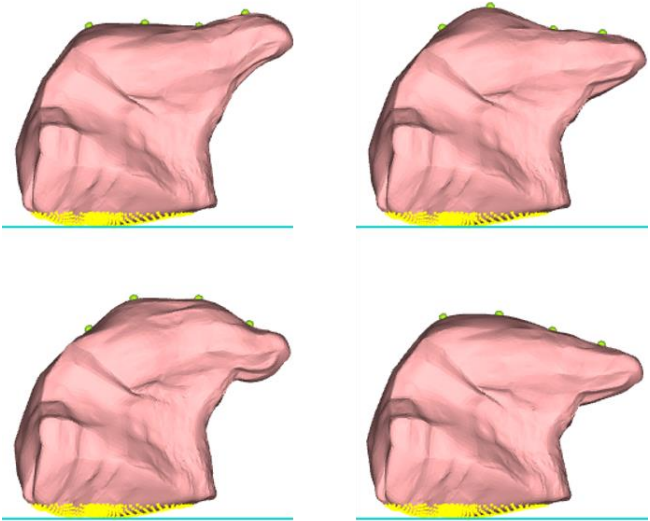

Fig. 2. Sample frames in the 3D tongue shape dataset.

**Step 3: Contour Extraction.** A modified active contour model (Snake model)-based method [19] is used to extract the contour in the ultrasound tongue image (as shown in Fig. 1(b)).

**Step 4: Similarity Measurement.** A measurement is made of the similarity between the contour extracted from the ultrasound image and the 2D contours projected from the 3D tongue shapes in the database. The definition of the similarity error is the mean sum of distances (MSD), which is widely used to measure the similarity between two curves; the smaller the MSD error is, the better the similarity. The detailed definition of MSD is given as follows:

$$\text{MSD}(V_1,V_2) = \frac{1}{2n}\left(\sum_{i=1}^{n}\min\left\|v_i^1 - v_j^2\right\| + \sum_{i=1}^{n}\min\left\|v_i^2 - v_j^1\right\|\right) \quad (4)$$

where $V_1$ is the contour extracted from the image and $V_2$ is the contour extracted from the 3D tongue shape in the database, $v_i^1$, $v_i^2$ are the elements of the contour $V_1$ and $V_2$ respectively. Here $n$ is the number of the elements of the contours (In our experiment, $n = 12$). Four constraint points generate $V_2$, while 12 points are selected to represent $V_1$. Consequently, to make the MSD measurement feasible, $V_2$ is re sampled equidistantly to keep the number of elements in the two contours the same.

During simulations, very small distances between constraint points were found to generate pathological curves. To retain smoothness in the tongue model, a penalty term was therefore added to the MSD error, defined as follows:

$$P = \sum_{i=2}^{m}\left(\frac{1}{\left\|v_i^2 - v_{i-1}^2\right\|}\right) \quad (5)$$

where $m$ is the number of constraint nodes before re-sampling (here $m$ is set as 4) and $v_i^2$ is the $i$ th constraint node. The overall objective function is now given as:

$$l = \alpha\left(1/\text{MSD}(V_1,V_2)\right) + \beta \times (1/P) \quad (6)$$

where $\alpha$ and $\beta$ are the weighting parameters (in our experiment, $\alpha = 0.8$ and $\beta = 0.2$).

In each time-step, this contour-based 3D deformation problem is implemented to measure the similarity of the contour extracted from 2D image and the contours projected from the 3D tongue shape. The most similar 3D tongue shape (the biggest $l$) will be selected to represent the target curve shape associated with the ultrasound frame.

The key reason for selecting the contour similarity measurement to create an association between the 2D ultrasound image and 3D tongue model is that, compared to ultrasound image similarity measurements or other similarity measurements using a 3D tongue model, measuring the similarity between 2D curves is of high efficiency. At the same time, although motion feature extraction from ultrasound tongue image still has difficulties, the contour extraction method is fairly robust in comparison with tissue points tracking method (or speckle tracking).

## 5. RESULTS

We implemented the aforementioned contour-based tongue motion system using Microsoft Visual C++ 2010 and MATLAB 2015a on a Windows 8 desktop computer with an Intel i7 3.7 GHz CPU and 16 GB RAM. The most time consuming step in our framework is the construction of the 3D tongue shape database, which was completed offline. The average processing time to build the association between current ultrasound frame and the 3D tongue model is about 1.2 seconds on our platform.

Here we select only four constraint nodes to drive the motion of the tongue on the 3D model's surface. In fact, the displacements of the constraint nodes must in reality be coupled since the tongue is a muscle-activated organ. However, the couple-relation is difficult to model. The compromise here is to use only four nodes to drive the model, with each node regarded as being independent of the others. Nevertheless, the deformation simulated with the proposed framework is informative and qualitatively realistic. Fig. 3 presents some results of the visualization platform on different vocalizations.

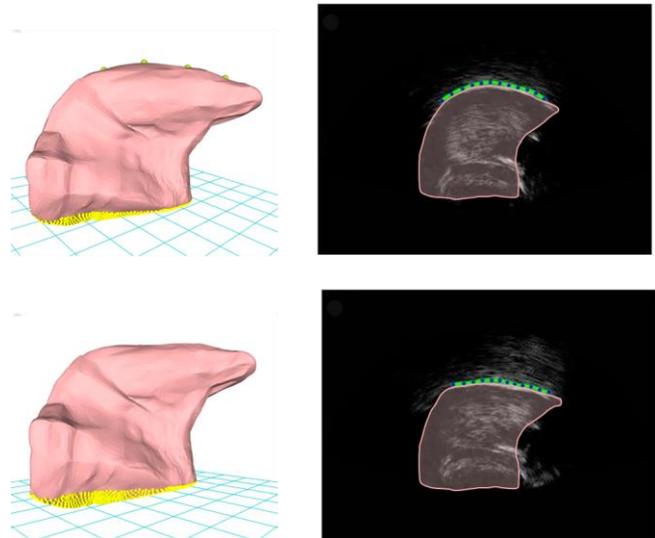

Fig. 4. Validation for the proposed method for 3D tongue modeling. The left row gives the 3D tongue model, while the right row gives the ultrasound tongue image with tongue extracted (the green lines donates the contour extracted). The midsagittal planes of the 3D tongue model are placed over the ultrasound tongue images in transparency.

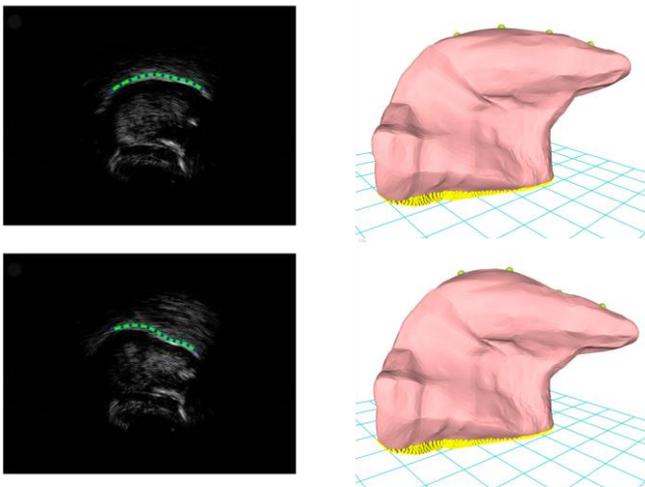

Fig. 3. Sample frames of 3D tongue modeling. The ultrasound images are given in the left column. The meaning of the color line and points is the same as Fig. 1. The 3D tongue shapes are given in the right column, which are selected from the 3D tongue database based on the method proposed in section 4.

As there is no effective quantitative evaluation method for the 3D tongue motion visualization presently, to further demonstrate the feasibility of the proposed method, the midsagittal plane of the 3D tongue model can be extracted from the model after the deformation. If the midsagittal contour of the 3D model can be fit to the ultrasound image, the effectiveness of the method will be validated. Fig. 4 gives some sample results, which demonstrate performance by visual observation.

## 6. DISCUSSION AND FUTURE WORK

In this paper, we briefly describe a novel contour-based 3D tongue motion visualization framework. The framework can be divided into three main parts: 1) 3D tongue shapes database construction; 2) Contour extraction from the B-mode ultrasound tongue image; 3) Similarity measurement (the definition is given in (6)) between the contour extracted from 2D ultrasound image and the contours projected from the 3D tongue shapes. The experiments conducted in section 5 demonstrate the promising potential applications of the proposed method in a variety fields.

There are still a number of improvements that can be brought to our present work. Firstly, the MSD error measurement may not be the optimal choice to measure the similarity between curves, and more specified measurement may need to be developed. Furthermore, there are non-midsagittal motions (or out-plane motions) of the tongue, and employing the motion information from midsagittal plane only is not enough to generate fully accurate tongue shapes. Lastly, the performance of the tongue motion visualization framework still needs to be evaluated quantitatively by making use of other imaging modality such as MRI and EMA.